\documentclass{article}

\PassOptionsToPackage{numbers, compress}{natbib}
\usepackage[final]{neurips_2024}

\usepackage[utf8]{inputenc} % allow utf-8 input
\usepackage[T1]{fontenc}    % use 8-bit T1 fonts
\usepackage{url}            % simple URL typesetting
\usepackage{booktabs}       % professional-quality tables
\usepackage{amsfonts}       % blackboard math symbols
\usepackage{nicefrac}       % compact symbols for 1/2, etc.
\usepackage{microtype}      % microtypography
\usepackage{xcolor}         % colors
\usepackage{makecell}
\usepackage{amsmath}

\usepackage{url}
\usepackage{booktabs}
\usepackage{multirow}
\usepackage{amsthm}
\newtheorem{definition}{Definition}

\usepackage{tabularx}
\usepackage{longtable}
\usepackage{array}
\usepackage{wrapfig}

\usepackage{amsmath,amsfonts}
\usepackage{booktabs}
\usepackage{rotating} % For rotating table
\usepackage{lipsum} % For generating filler text
\usepackage{array}
\usepackage{multirow}
\usepackage[most]{tcolorbox}

\usepackage{graphicx}

\usepackage{caption}
\usepackage{subcaption}
\usepackage{pifont}% http://ctan.org/pkg/pifont
\newcommand{\cmark}{\ding{51}}%
\definecolor{lightgray}{gray}{0.75}
\newcommand{\xmark}{\textcolor{lightgray}{\ding{55}}}% X mark symbol
\definecolor{economist}{RGB}{115,00,00} % Adjust the RGB values to get the desired shade
\definecolor{customgreen}{RGB}{116, 154, 114}
\definecolor{lightgreen}{RGB}{240, 246, 232}
\definecolor{greylight}{RGB}{242, 242, 242}
\definecolor{greydark}{RGB}{179, 179, 179}
\definecolor{ForestGreen}{RGB}{34, 139, 34}
\definecolor{pastelblue}{RGB}{46, 95, 127} % Light Blue
\definecolor{pastelorange}{RGB}{201, 171, 102} % Light Orange
\definecolor{pastelgreen}{RGB}{76, 124, 49} % Light Orange
\newtcolorbox{greycustomblock}{
  colframe=greydark,        % Left line color
  colback=greylight,        % Background color
  boxrule=1pt,              % Line width
  left=2.5pt,               % Inner left margin
  right=3pt,                % Inner right margin
  top=5pt,                  % Inner top margin
  bottom=3pt,               % Inner bottom margin
  arc=0pt,                  % No rounded corners
  breakable,                % Allow breaking across pages
  before skip=0.2\baselineskip, % Vertical space before
  after skip=0.2\baselineskip,  % Vertical space after
  left skip=0pt,            % Left skip (adjust as needed)
  right skip=0pt,           % Right skip (adjust as needed)
  enhanced jigsaw,          % Enhanced jigsaw for precise control
  frame hidden,             % Hide the default frame
  overlay={                 % Custom overlay for left line
    \draw[greydark, line width=2pt]
      ([yshift=-1pt]frame.north west) -- ([yshift=1pt]frame.south west); % Adjusting the y coordinates to match exactly
  },
  fontupper=\fontfamily{lmr}\selectfont, % Font size and style for the table
}
\usepackage{fontawesome}
\usepackage{adjustbox} 
\usepackage{tikz}
\usepackage{hyperref}       % hyperlinks

\usepackage[framemethod=TikZ]{mdframed}
\mdfsetup{% 
    middlelinecolor =   none,
    middlelinewidth =   1pt,
    backgroundcolor =   blue!3,
    roundcorner     =   5pt,
}

\newcommand{\lightbulbicon}{%
  \begin{tikzpicture}[baseline=-0.5ex]
    \draw[fill=white, draw=customgreen, thick] (0,0) circle (1.5ex);
    \node[scale=0.8, color=customgreen] at (0,0) {\faLightbulbO~};
  \end{tikzpicture}%
}

\newtcolorbox{customblockquote}{
  colframe=customgreen,
  colback=lightgreen,
  boxrule=0pt,
  leftrule=2pt, 
  left=1pt,  % Set to 0pt so the background color touches the left line
  right=3pt,
  top=5pt,
  bottom=3pt,
  arc=0pt,
  breakable,
  before skip=1.1\baselineskip,
  after skip=0.7\baselineskip,
  left skip=0pt,
  right skip=0pt,
  enhanced jigsaw,
  frame hidden,
   overlay={
    \draw[customgreen, line width=2pt] 
      (frame.north west) -- (frame.south west);
    \node[inner sep=0pt] at ([xshift=0pt, yshift=-1.3pt]frame.north west) {\lightbulbicon};
  },
  fontupper=\fontfamily{lmr}\selectfont,
  boxsep=3pt,
}

\usepackage{listings}
\definecolor{codegreen}{rgb}{0,0.6,0}
\definecolor{codegray}{rgb}{0.5,0.5,0.5}
\definecolor{codepurple}{rgb}{0.58,0,0.82}
\definecolor{backcolour}{rgb}{0.95,0.95,0.92}
\lstdefinestyle{mystyle}{
    backgroundcolor=\color{backcolour},   
    commentstyle=\color{codegreen},
    keywordstyle=\color{magenta},
    numberstyle=\tiny\color{codegray},
    stringstyle=\color{codepurple},
    basicstyle=\ttfamily\footnotesize,
    breakatwhitespace=false,         
    breaklines=true,                 
    captionpos=b,                    
    keepspaces=true,                 
    numbers=left,                    
    numbersep=5pt,                  
    showspaces=false,                
    showstringspaces=false,
    showtabs=false,                  
    tabsize=2
}
\usepackage[utf8]{inputenc}
\usepackage[english]{babel}
\usepackage{amssymb,amsmath,amsthm}
\usepackage{xcolor}
\usepackage{algorithm}
\usepackage{tikz}
\usepackage{algpseudocode}
\usepackage{multirow}

\newcommand{\mycomment}[1]{}

\usepackage{mathtools} % Bonus
\title{Quantifying perturbation impacts for large language models}

\author{%
  Paulius Rauba\thanks{Equal Contribution} \\
  University of Cambridge\\
  \texttt{pr501@cam.ac.uk} 
  \And
  Qiyao Wei$^*$ \\
  University of Cambridge\\
  \texttt{qw281@cam.ac.uk} 
  \And
  Mihaela van der Schaar \\
  University of Cambridge\\
  \texttt{mv472@cam.ac.uk}
}

\begin{document}

\maketitle

\begin{abstract}
We consider the problem of quantifying how an input perturbation impacts the outputs of large language models (LLMs), a fundamental task for model reliability and post-hoc interpretability. A key obstacle in this domain is disentangling the meaningful changes in model responses from the intrinsic stochasticity of LLM outputs. To overcome this, we introduce \textit{Distribution-Based Perturbation Analysis} (DBPA), a framework that reformulates LLM perturbation analysis as a frequentist hypothesis testing problem. DBPA constructs empirical null and alternative output distributions within a low-dimensional semantic similarity space via Monte Carlo sampling. Comparisons of Monte Carlo estimates in the reduced dimensionality space enables tractable frequentist inference without relying on restrictive distributional assumptions. The framework is model-agnostic, supports the evaluation of arbitrary input perturbations on any black-box LLM, yields interpretable p-values, supports multiple perturbation testing via controlled error rates,  and provides scalar effect sizes for any chosen similarity or distance metric. We demonstrate the effectiveness of DBPA in evaluating perturbation impacts, showing its versatility for perturbation analysis.
\end{abstract}

\section{Introduction}
Large language models (LLMs) generate outputs conditioned on textual inputs by iteratively sampling from a distribution of tokens. Therefore, the outputs of LLMs exhibit inherent variability due to the stochastic sampling process, a process controlled via parameters such as temperature or top-k \citep{renze2024effect}. This means that evaluating how a specific input perturbation---changing some information in the prompt---affects the output, is not straightforward \cite{romero2024resilient}. Understanding and quantifying the effects of such perturbations is crucial in high-stakes applications—such as legal document drafting or medical diagnosis—where errors or unintended behavior could have significant consequences \cite{mesko2023imperative, helberger2023chatgpt}.

Systematic evaluation of output responses to input perturbation is fundamental to comprehending LLM behavior. It provides quantitative insights into model robustness and output consistency across diverse input conditions. Perturbation analysis can serve at least three crucial functions in general ML models. First, it helps with \textit{vulnerability identification} by quantifying potential vulnerabilities to adversarial attacks \cite{goodfellow2014explaining}. Second, it aids with \textit{bias discovery}, whereby latent biases or unintended behaviors may not become apparent through single-dimension auditing approaches \cite{ribeiro2016should}. Third, it can work within a \textit{compliance framework}. Measurable frameworks for assessing model behavior are essential for compliance with emerging ethical and legal accountability regulations \cite{doshi2017accountability}. In the context of language models, subtle changes in patient history could lead to wildly different diagnoses; and patients with nearly identical health records might receive drastically different treatment plans due to minor grammatical changes. Given these critical functions, there is a clear need for a comprehensive LLM auditing framework centered around perturbation analysis. 

Current methods for analyzing the impact of input perturbations often focus on simplistic metrics, such as word overlap or direct log-probability comparisons. While effective in certain cases, these approaches fail to account for the nuanced, high-dimensional nature of semantic information processed by LLMs \cite{bhandari2020re, lin2004rouge}. Moreover, existing methods typically lack rigorous statistical foundations, making it difficult to disentangle meaningful changes in model behavior from intrinsic randomness in the output generation process. Efforts to address this have included specialized attribution methods, feature importance techniques, and counterfactual reasoning frameworks \cite{ribeiro2016should, garg2019counterfactual}. However, these approaches are often model-specific, rely on restrictive assumptions about the data or model, or fail to provide interpretable and generalizable metrics\footnote{Code available at https://github.com/vanderschaarlab/DBPA}.

\textbf{Our solution}. In this work, we introduce Distribution-Based Perturbation Analysis (DBPA), a framework that reformulates the problem of LLM perturbation analysis as a frequentist hypothesis testing task. DBPA constructs empirical output distributions using Monte Carlo sampling to capture the inherent stochasticity of LLMs, and evaluates perturbation effects within a low-dimensional semantic similarity space. By leveraging statistical hypothesis testing, the framework enables robust, interpretable inferences about whether and how input perturbations meaningfully influence LLM outputs. DBPA is model-agnostic, computationally efficient, and flexible enough to accommodate arbitrary perturbations on any black-box LLM. It also provides interpretable p-values, scalar effect sizes, and supports multiple testing with controlled error rates, making it a versatile tool for post-hoc interpretability and reliability assessments of LLMs.

\begin{table*}[t]
\centering
\caption{\textbf{Use case examples of distribution-based perturbation analysis}. We show how DBPA can be used in five areas for different tasks with the associated added value of our framework.}
\tiny
\label{tab:DBPA-summary}
\resizebox{\textwidth}{!}{%
\begin{tabular}{@{}p{0.13\textwidth}p{0.25\textwidth}p{0.25\textwidth}p{0.25\textwidth}@{}}
\toprule
\textbf{Area} & \textbf{Description of task} & \textbf{Use case of DBPA} & \textbf{Value added} \\
\midrule
Prompt robustness & Evaluate sensitivity to question rephrasing in QA systems & Ensure consistent answers across question formulations & Improved system reliability \\
\addlinespace
Training Stability & Evaluate output consistency across fine-tuning iterations & Monitor and optimize model reliability over time & Better training stability monitoring \\
\addlinespace
Model comparison & Analyze the impact of different LLM answers to reference answers (e.g. GPT-4) & Determine optimal model & Best model chosen \\
Adversarial attack & Evaluate whether a model is being adversarially attacked & Determine if the output diverges too much from relevant, seen answers & Better adversarial robustness \\ 
\bottomrule
\end{tabular}%
}
\end{table*}

\textbf{Significance beyond technical novelty}. We see this work as having immediate practical relevance for practitioners who wish to evaluate their LLM outputs and whether they change in real-world practical settings, including evaluating prompt robustness, LLM robustness, and conducting model comparisons. 

\begin{customblockquote}
    \textbf{Contributions}. \textbf{\textcolor{pastelgreen}{\textcircled{1}}} We identify limitations in existing sensitivity-based measures for language models (Sec. \ref{sec:limitations}).  \textbf{\textcolor{pastelgreen}{\textcircled{2}}} We introduce distribution-based perturbation analysis which is a model-agnostic sensitivity technique that can test the effect of \textit{any perturbation} with statistical significance measures (Sec. \ref{sec:sensitivity}). \textbf{\textcolor{pastelgreen}{\textcircled{3}}} We perform multiple case studies to show the usefulness of DBPA (Sec. \ref{sec:case_studies}).
\end{customblockquote}

\vspace{-2mm}
\section{An analysis of viewing LLM outputs through frequentist hypothesis testing}
\label{sec:limitations}

In this section, we describe how we can view LLM outputs through the lens of frequentist hypothesis testing. We will first define the problem, then pose as hypothesis testing, and discuss why naively applying statistical methods will fail due to curse of dimensionality. 

\subsection{Problem formulation}

Let $\mathcal{X}$ denote the input space and $\mathcal{Y}$ the output space of a machine learning system. We define the system as a stochastic mapping $\mathcal{S}: \mathcal{X} \rightarrow \mathcal{P}(\mathcal{Y})$, where $\mathcal{P}(\mathcal{Y})$ is the space of probability distributions over $\mathcal{Y}$. This formulation captures the inherent stochasticity of modern ML systems, including LLMs.

Our objective is to address the following research question: Given an ML system $\mathcal{S}$, an input $x \in \mathcal{X}$, and a perturbation $\Delta$, how can we systematically measure and interpret the impact of $\Delta$ on the output distribution of $\mathcal{S}(x)$ under a general notion of sensitivity?

\begin{definition}[Sensitivity]
\label{def:sensitivity}

The sensitivity of a machine learning system $\mathcal{S}$ with respect to a perturbation $\Delta: \mathcal{X} \rightarrow \mathcal{X}$, or a system perturbation $\Delta: \mathcal{S} \rightarrow \mathcal{S}$, at input $x \in \mathcal{X}$, is characterized by a measure of discrepancy between the output distributions $\mathcal{S}(x)$ and $\mathcal{S}(\Delta x)$, or $\mathcal{S}(x)$ and $\Delta \mathcal{S}(x)$, respectively.
\end{definition}

This definition encompasses both input perturbations ($\Delta \mathcal{Y} == \mathcal{S}(\Delta x)$) and system perturbations ($\Delta \mathcal{Y} == \Delta \mathcal{S}(X)$). Crucially, we wish to evaluate this sensitivity by only having access to the ML system $\mathcal{S}$ from which we could sample. We do not assume access to ground-truth labels and make no assumptions on the architecture of $\mathcal{S}$.

\subsection{Distribution testing as a frequentist hypothesis problem}
The challenge of assessing the sensitivity of $\mathcal{S}$ can be reframed as a frequentist hypothesis testing problem. Traditionally, one might be tempted to compare individual outputs of $\mathcal{S}$ for different inputs:

\begin{equation}
    y = \mathcal{S}(x), \quad y' = \mathcal{S}(\Delta(x))
\end{equation}

However, this approach is fundamentally flawed due to the stochastic nature of LLM systems. For any fixed input $x$, $\mathcal{S}(x)$ is a random variable, and thus $y$ and $y'$ are single realizations from a distribution of outputs. Consequently, any observed difference between $y$ and $y'$ could be due to inherent randomness rather than a true effect of $\Delta$.

To address this limitation, we propose reframing the problem from the lens of \textit{distribution testing}. Instead of comparing individual outputs, we compare the entire output distributions. Let $\mathcal{D}_x$ and $\mathcal{D}_{\Delta(x)}$ be the distributions of outputs from $\mathcal{S}(x)$ and $\mathcal{S}(\Delta(x))$ respectively. 

\begin{definition}[Output Distribution]
For a given input $x \in \mathcal{X}$ and an LLM $\mathcal{S}: \mathcal{X} \rightarrow \mathcal{P}(\mathcal{Y})$, the output distribution $\mathcal{D}_x$ is a probability distribution over the output space $\mathcal{Y}$ such that:

\begin{equation}
\mathcal{D}_x(y) = \mathbb{P}(\mathcal{S}(x) = y), \quad \forall y \in \mathcal{Y}
\end{equation}
\end{definition}

Our goal is to determine whether these distributions differ significantly. This can be formulated as a hypothesis test:

\begin{align}
    H_0&: \mathcal{D}_x = \mathcal{D}_{\Delta(x)} \quad \text{(The perturbation has no effect)} \\
    H_1&: \mathcal{D}_x \neq \mathcal{D}_{\Delta(x)} \quad \text{(The perturbation affects the output distribution)}
\end{align}

The primary benefit of such a distributional formulation is that it captures the full stochastic behavior of $\mathcal{S}$ instead of just a single realization. This means we could perform statistical inference by directly comparing these distributions and understanding how much the outputs have shifted across the whole output space, even detecting subtle shifts that might not be apparent from individual samples.

%\textbf{Benefits of a distributional formulation}. Formulating perturbation analysis as distribution comparison offers several advantages: (i) it captures the full stochastic behavior of $\mathcal{S}$, not just a single realization; (ii) it allows for statistical inference, including p-values and effect sizes; (iii) it can detect subtle shifts that might not be apparent from individual samples; (iv) Formulating perturbation analysis through the lens of distribution testing has multiple benefits and is generalizable to any perturbation $\Delta$ and any stochastic language model $\mathcal{S}$, providing a universal framework for perturbation analysis.

\begin{customblockquote}
To understand the sensitivity of a language model to changes in input, we need to look at the entire range of possible outputs it can produce, not just single examples.
\end{customblockquote}

However, while theoretically grounded, the usage of such a hypothesis testing framework in the context of LLMs poses unique practical challenges that do not appear in regular settings. We discuss these next.

\subsection{Challenges with analyzing output distributions}
There are two primary challenges in comparing output distributions to evaluate the effect of an input perturbation on the output: computational intractability and poor interpretability.

$\blacktriangleright$ \textbf{Challenge 1: Computational intractability}. In almost all cases, it is impossible to directly evaluate the output distribution change because of the exponential output space $\mathcal{Y}$ = $V^L$ where $V$ is the vocabulary and $L$ is the sequence length, yielding $|Y| = |V|^L$ possible outputs for each sentence, and that is assuming a fixed $L$. Suppose we limit our language model $\mathcal{S}$ to output only sequences of length $L$. In this case, $\mathcal{D}_x$ would require summing over all possible sequences:

\begin{equation}
    \mathcal{D}_x = \sum_{y_1 \in V} \sum_{y_2 \in V} \cdots \sum_{y_L \in V} \prod_{t=1}^L p(y_t|y_{<t}, x) \cdot \delta_{(y_1, \ldots, y_L)}
\end{equation}

where $p(y_t|y_{<t}, x)$ is the probability of token $y_t$ given the previous tokens and input $x$, and $\delta_{y_1, \ldots, y_L}$ is the Dirac delta function assigning the probability to the specific sequence. Even in such a case, this would be an intractable computation for any existing operating system with exponential computational complexity $O(|V|^L)$.

$\blacktriangleright$ \textbf{Challenge 2: Interpretability}. A second major issue is that the distribution $\mathcal{D}_x$ does not provide an interpretable understanding of the LLM output. As LLMs are increasingly being employed as reasoning engines \cite{yao2022react, hao2023reasoning}, we care about whether their outputs differ semantically, not probabilistically. For instance, suppose $\mathcal{S}$ outputs two answers to a question on treatment recommendations

$y_1$: ``\textit{Targeted radiation therapy is suggested}'', $y_2$: ``\textit{We suggest targeted radiation therapy}''.

Probabilistically, the answer distributions do not match, i.e. $\mathcal{D}_x(y_1) \neq \mathcal{D}_x(y_2)$ in general. Semantically however, we see them as having the same inherent recommendation.

Ideally, we'd like to be able to resolve both issues at the same time --- (i) be able to computationally approximate the distribution \textit{and} (ii) evaluate whether the differences are semantically meaningful, not just probabilistically different. We show that we can achieve both with finite-sample approximations.

\begin{customblockquote}
Analyzing output distributions of language models faces two unique challenges: the computational intractability due to the enormous output space, and the need for semantic rather than just probabilistic interpretation of differences.
\end{customblockquote}

\subsection{Finite sample approximation to output distributions}

We have established that using distributions to analyze LLM outputs offer significant advantages compared to using a single output. These advantages come with two practical challenges: (i) computational intractability and (ii) poor interpretability. Here, we suggest that using finite sample approximations can resolve both challenges.

\begin{wrapfigure}[14]{r}{0.5\columnwidth}
    \vspace*{-5mm}
    \centering
    \includegraphics[width=\linewidth]{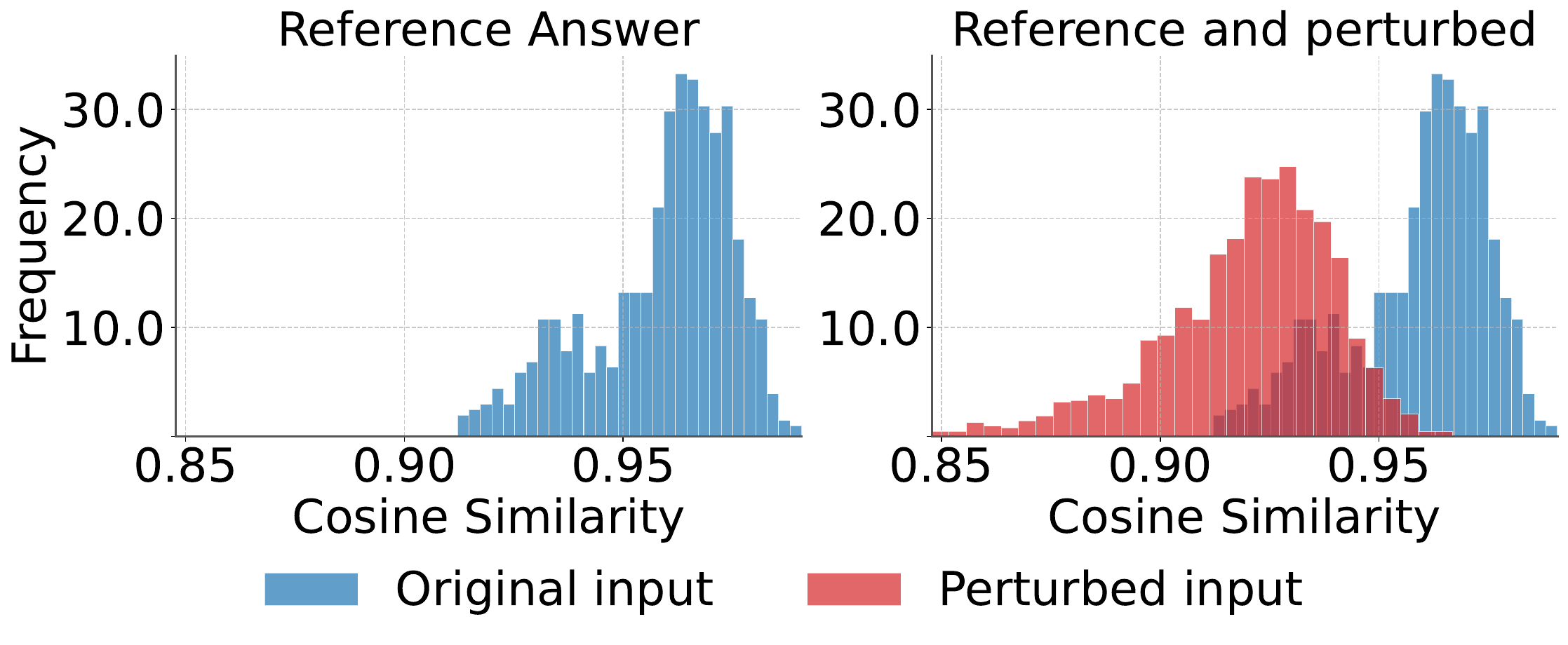}
    \caption{\textbf{Example of null and alternative distributions.} The null distribution $P_0$ (left, blue) is constructed based on the intrinsic variability of responses. The alternative distribution with a perturbed input $P_1$ (right, red) is quantified with respect to the original distributions. This measures the output distribution change given a perturbed prompt in the cosine similarity space.}
    \label{fig:null_distribution}
    \vspace{-2mm}
    \rule{\linewidth}{.5pt}
\end{wrapfigure}

$\blacktriangleright$ \textbf{Addressing challenge 1: Computational complexity}. We address the computational complexity is by Monte Carlo sampling. We define the stochastic approximations of the output distributions for an input $x$ and its perturbation $\Delta(x)$ as:

\begin{align}
\hat{\mathcal{D}}_x &= \{y_i\}_{i=1}^k, \quad y_i \stackrel{i.i.d.}{\sim} \mathcal{S}(x) \\
\hat{\mathcal{D}}_{\Delta(x)} &= \{y'_i\}_{i=1}^k, \quad y'_i \stackrel{i.i.d.}{\sim} \mathcal{S}(\Delta(x))
\end{align}

where $k$ is the sample size. This simply samples $k$ outputs for the original and perturbed inputs. 

$\blacktriangleright$ \textbf{Addressing challenge 2: Interpretability}. Given a finite set of samples, we would like to measure how much the output varies given an input perturbation. To quantify the distributional changes induced by input perturbations, we introduce the similarity metric $s: \mathcal{Y} \times \mathcal{Y} \rightarrow [0, 1]$ (discussed more in Sec. \ref{sec:design}). This metric allows us to construct empirical distributions of pairwise similarities:

\begin{align}
P_0 &= \{s(y_i, y_j) : y_i, y_j \in \hat{\mathcal{D}}_x, i \neq j\} \\
P_1 &= \{s(y_i, y'_j) : y_i \in \hat{\mathcal{D}}_x, y'_j \in \hat{\mathcal{D}}_{\Delta(x)}\}
\end{align}

Here, $P_0$ captures the intrinsic variability within the original output (equivalent to the \textit{null distribution}), whereas $P_1$ captures the cross-distribution similarities between the original and perturbed outputs (equivalent to the \textit{alternative distribution}), c.f. Figure \ref{fig:null_distribution}. We have therefore constructed two distributions which represent the variability in answer similarities as a proxy for sensitivity (Def. \ref{def:sensitivity}). In Sec. \ref{sec:sensitivity}, we show how such distributions can be used to obtain sensitivities of perturbations and associated p-values.

$\blacktriangleright$ \textbf{Advantages of such a formulation}. With this formulation, we (i) capture the stochastic nature of LLM outputs more faithfully than point estimates; (ii) connect LLM outputs to frequentist hypothesis testing; (iii) quantify effect size; (iv) maintain model and input agnosticism.

In the next section, we provide an end-to-end illustration of DBPA for LLM outputs.

\begin{customblockquote}
Finite sample approximations using Monte Carlo sampling and pairwise similarity metrics offer a computationally feasible and interpretable approach to analyzing output distributions of language models. 
\end{customblockquote}

\section{Distribution-based perturbation analysis}
\label{sec:sensitivity}

We present a novel model-agnostic methodology for assessing the sensitivity of LLMs to perturbations. Our approach avoids restrictive distributional assumptions and utilizes the entire output distribution of $\mathcal{S}$, capturing the intrinsic variability in LLM responses. We enable frequentist statistical hypothesis testing using p-values through the construction of null and alternative distributions. Importantly, our framework is applicable to \textit{any perturbation} and \textit{any language model}, with the minimal requirement of being able to sample from the language model's output distribution and construct embeddings.

\subsection{The procedure}

\textbf{Objective}. DBPA aims to evaluate two values. First, we calculate the \textit{ ffect size}(``by how much has the response distribution shifted``?). Second, we calculate the \textit{p-value} (``is the difference between the new and old distributions significant``?). DBPA follows a simple and intuitive four-step procedure, the general form of which is outlined below.
\begin{greycustomblock}
\textbf{distribution-based perturbation analysis in a nutshell.} 

\vspace{-2.5mm}
\rule{\linewidth}{.5pt}
\vspace{1mm}

distribution-based perturbation analysis for language models consists of four main steps: response sampling, distribution construction, comparison, and statistical inference.
\vspace{1mm}

\textbf{I. Response Sampling}. We sample $k$ i.i.d. responses from both the original and perturbed input distributions:
\begin{align*}
\hat{\mathcal{D}}_x &= \{y_1, \ldots, y_k\}, \quad y_i \stackrel{i.i.d.}{\sim} \mathcal{S}(x) \\
\hat{\mathcal{D}}_{\Delta(x)} &= \{y'_1, \ldots, y'_k\}, \quad y'_i \stackrel{i.i.d.}{\sim} \mathcal{S}(\Delta(x))
\end{align*}
where $\mathcal{S}$ is the language model, $x$ is the original input, and $\Delta(x)$ is the perturbed input.
\vspace{1mm}

\textbf{II. Distribution construction}. We construct empirical null and alternative distributions using a similarity metric $s: \mathcal{Y} \times \mathcal{Y} \rightarrow [0, 1]$:
\begin{align*}
P_0 &= \{s(y_i, y_j) : y_i, y_j \in \hat{\mathcal{D}}_x, i \neq j\} \\
P_1 &= \{s(y_i, y'_j) : y_i \in \hat{\mathcal{D}}_x, y'_j \in \hat{\mathcal{D}}_{\Delta(x)}\}
\end{align*}
where $s$ is an appropriate similarity-preserving and dimensionality-reducing metric, such as the cosine distance. 

\vspace{1mm}
\textbf{III. Distributional Comparison}. We quantify the discrepancy between the null and alternative distributions using a general metric $\omega$:
\begin{align*}
\omega(P_0, P_1) : \mathcal{P} \times \mathcal{P} \rightarrow \mathbb{R}_{\geq 0}
\end{align*}
where $\mathcal{P}$ is the space of probability distributions and $\omega$ measures the difference between two distributions.
\vspace{1mm}

\textbf{IV. Statistical Inference}. We formulate our perturbation analysis as a hypothesis test:
\begin{align*}
H_0&: \mathcal{S}(x) = \mathcal{S}(\Delta(x)) \\
H_1&: \mathcal{S}(x) \neq \mathcal{S}(\Delta(x))
\end{align*}
To assess statistical significance, we employ a permutation test. This approach allows us to compute p-values while only assuming exchangeability \textit{in the similarity space}.
\vspace{1mm}

\textbf{Objective}. The goal of distribution-based perturbation analysis is to determine whether there is a statistically significant difference between the output distributions of the original and perturbed inputs and its associated effect size. We observe the effect size as $\omega$. To obtain a frequentist p-value, we compare the observed $\omega(P_0, P_1)$ to the distribution of $\omega$ values obtained through permutation, yielding a p-value:
\begin{equation}
p = \mathbb{P}(\omega(P_0^*, P_1^*) \geq \omega(P_0, P_1) | H_0)
\end{equation}
where $P_0^*$ and $P_1^*$ are permuted versions of $P_0$ and $P_1$ under the null hypothesis.
\end{greycustomblock}

%-> embedding function must be solid (i.e. non-random embeddings) -> explain. discussion on the embedding and what would happen if we used the LLM embeddings

\subsection{Implementation essentials}
\label{sec:design}
There are a few practical implementation essentials to take into account when developing DBPA. 

\textbf{Why use scalar pairwise cosine similarities instead of directly using high-dimensional emebddings?} Using non-reduced embeddings to construct null and alternative distributions faces two significant challenges. $\blacktriangleright$  \textbf{High dimensionality}. Embedding spaces typically have hundreds or thousands of dimensions, making direct distribution estimation in this space problematic due to the curse of dimensionality. $\blacktriangleright$ \textbf{Excess semantic information}. Embeddings capture rich semantic information, much of which is irrelevant since our only focus is to understand how much the responses vary in response to changes in x, as opposed to their inherent semantic position in the embedding space. 

\begin{wrapfigure}[8]{r}{0.5\columnwidth}
    \vspace*{-8mm}
    \centering
    \includegraphics[width=\linewidth]{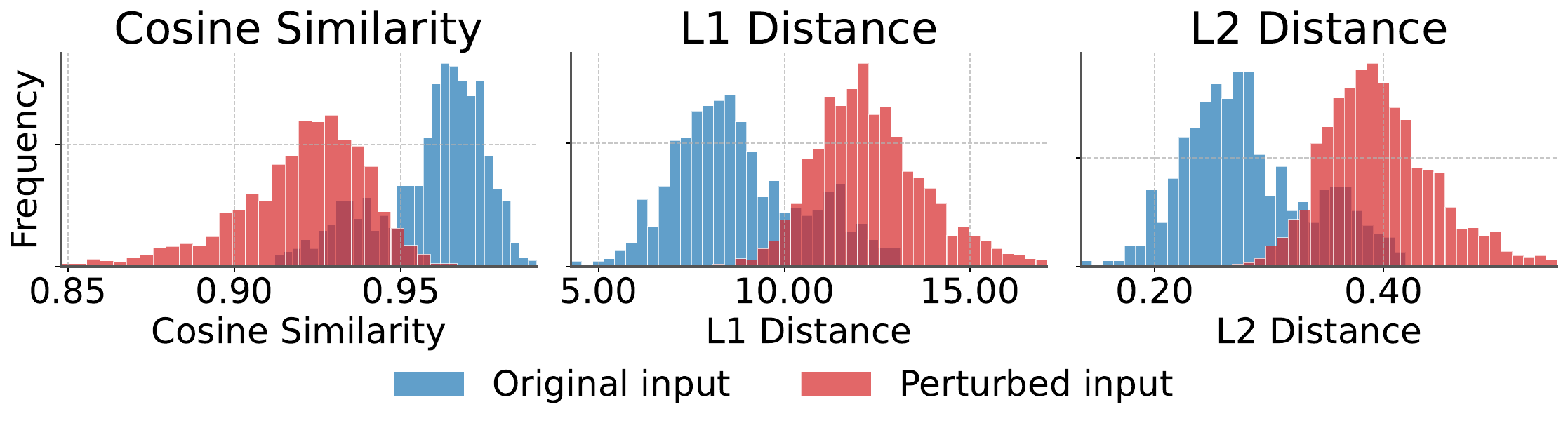}
    \caption{\textbf{Examples of different dimensionality metrics.} We show that we can use multiple metrics to reduce the dimensionality of given embeddings. }
    \label{fig:similarity_metrics}
    \vspace{-2mm}
    \rule{\linewidth}{.5pt}
\end{wrapfigure}

\textbf{What is an appropriate metric to overcome the challenges with high-dimensional and semantically-rich embeddings?} We desire to employ a similarity-preserving dimensionality reduction technique to project these distributions onto a tractable, low-dimensional manifold. While there are many metrics that have such properties, such as the L1 or L2 norms, we find the cosine similarity $    s(y_i, y_j) = \frac{\langle e(y_i), e(y_j) \rangle}{\|e(y_i)\| \|e(y_j)\|}$ to be a natural choice, where $e(\cdot)$ is an embedding function.

\textbf{What is an appropriate distance metric $\omega$?} The reason why there exists a choice for $\omega$ is that we are dealing with the comparison between two distributions. This is different from traditional resampling-based approches that construct a null distribution and evaluate a single instance against it \cite{yu2003resampling}. While the choice for $\omega$ might vary depending on the application, we employ the Jensen-Shannon divergence (JSD) as a measure for $\omega$: $\text{JSD}(P_0 \| P_1) = \frac{1}{2}\left(D_{\text{KL}}(P_0 \| M) + D_{\text{KL}}(P_1 \| M)\right)$, where $M = \frac{1}{2}(P_0 + P_1)$ and $D_{\text{KL}}$ is the Kullback-Leibler divergence. This is because JSD has three useful properties for evaluating distributions: (i) symmetry, ensuring that the measure is invariant to the order of the distributions being compared; (ii) boundedness, providing a consistent scale for interpretation across different inputs and perturbations; and (iii) sensitivity to differences in both the location and shape of the distributions. 

\textbf{Why perform permutation-based testing instead of directly MC-sampling from $\mathcal{S}$?} In fact, it is entirely possible to use MC-sampling to obtain p-values. However, this comes at a cost---directly sampling $S(x)$ and from $S(\Delta(x))$ might be computationally expensive, as this requires directly generating new samples from $\mathcal{S}$. In contrast, permutation-based testing gives us, in expectation, equivalent distributions at lower computational costs. We expand on this in Appendix \ref{appendix:mc_sampling}

\section{Case studies}
\label{sec:case_studies}

We demonstrate the effectiveness of our method on a variety of use cases. In the following subsections, we will show that our method can (1) capture those answer divergences that are significant and those that are not under perturbation (2) analyze the robustness of language models to irrelevant changes in the prompt (3) evaluate alignment with reference language model. By default, we run the experiment over 5 seeds, and report the mean and standard deviation of the measurements. We instantiate the DBPA metrics as the p-value and the effect size $\omega$, computed as the JSD distance between the null and alternative distributions.

\subsection{DBPA can measure answer divergence under perturbation}
\label{sec:role-play}

In this experiment, we show that the DBPA framework provides a way to measure how much the LLM is influenced by a perturbation in the prompt. If the LLM is influenced by the perturbation, we expect it to produce a different answer that is statistically significant. 

\begin{figure}[t]
    \centering
    \includegraphics[width=1.0\linewidth]{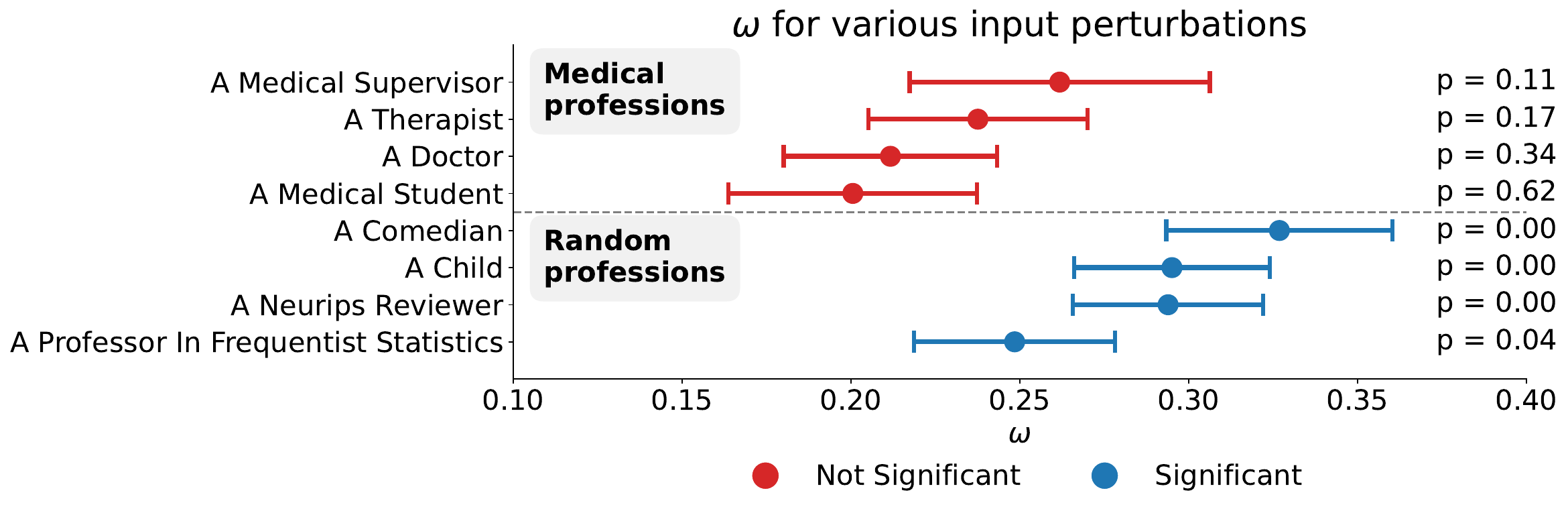}
    \caption{\textbf{Measuring the effect size $\omega$ and statistical significance of outputs when prefixing the original question with various "Act as..." prompts.} Results show that relevant professional roles (e.g., medical professions) yield consistent outputs, while diverse roles produce significantly different responses, demonstrating the framework's ability to quantify prompt perturbation effects. If $p < \alpha$, where $\alpha = 0.05$, we say that the output distribution is significant.}
    \label{fig:role-play}
\end{figure}

\textbf{Setup}. (1) We query the LLM a healthcare question, and use the LLM response as the null distribution. (2) We then perturb the prompt, and measure whether the LLM produces a different answer that is statistically significant. We perturb the prompt by asking the LLM to role-play. In front of the original question, we append the prompt "Act as ...", where "..." is labeled as the y-axis of Figure \ref{fig:role-play}. The LLM used in this experiment is gpt-3.5.

\textbf{Discussion}. We show the effect size $\omega$ and the p-value plotted against various "Act as ..." perturbations. Because we are asking the LLM medical questions, we expect the answer to not change when the LLM role-plays as any medical profession (top half of Figure \ref{fig:role-play}), and change significantly when the LLM role-plays as other diverse professions (bottom half of Figure \ref{fig:role-play}). We find that we are able to capture the LLM response variability. Whenever the repsonses are prepended to role-play in a way that significantly changes the output distribution (``Random professions''), this is captured by the $\omega$ and the p-value. In contrast, ``Medical professions'' do not produce statistically significant responses. \textbf{Takeaway}: The DBPA framework effectively quantifies how different prompt perturbations impact language model outputs.

\subsection{DBPA can measure the robustness of language models to irrelevant changes in the prompt}
\label{sec:propmt-robust}

\textbf{Setup}. This experiment aims to demonstrate how the DBPA framework can measure LLM robustness to prompt variations. We query an LLM with an original question, then with multiple paraphrased versions that maintain the same meaning. We then compute the p-values and $\omega$ for responses to paraphrased prompts. We calculate the proportion of responses showing significant changes (p-value < $\alpha$), where $\alpha=0.05$, and average $\omega$. This process is repeated for different LLMs, with results in Table \ref{tab:prompt-robust}. A robust model should show few significant changes and small effect sizes.

\begin{wraptable}[14]{r}{.48\linewidth}
    \vspace{-\baselineskip-1.8pt}
    \centering
    \caption{\textbf{Measuring P(p-value $< 0.05$) and effect size $\omega$ across LLMs}. In this context, P(p-value $< 0.05$) should be interpreted as computing how many statistically-significant shifts there are in the perturbed responses, i.e. how many responses have p-value $< 0.05$. As we expect, more powerful models tend to be more robust.}
    \vspace{-\baselineskip}
    \smallskip
    \label{tab:prompt-robust}
    \resizebox{\linewidth}{!}{%
    \begin{tabular}{@{}l|c|c@{}}
        \toprule
        \bf LLM & \bf\makecell{P(p-value $< 0.05$)} & \bf\makecell{$\omega$} \\
        \midrule
        gpt-4-0613-20231016 & 0.15 (0.3) & 0.23 (0.06) \\
        gpt-35-1106 & 0.05 (0.06) & 0.22 (0.04) \\
        microsoft/Phi-3-mini-4k-instruct & 0.1 (0.05) & 0.23 (0.03) \\
        openai-community/gpt2 & 0.18 (0.22) & 0.27 (0.08) \\
        meta-llama/Meta-Llama-3.1-8B-Instruct & 0.25 (0.11) & 0.27 (0.01) \\
        mistralai/Mistral-7B-Instruct-v0.2 & 0.3 (0.3) & 0.25 (0.05) \\
        google/gemma-2-9b-it & 0.33 (0.22) & 0.26 (0.07) \\
        HuggingFaceTB/SmolLM-135M & 0.35 (0.24) & 0.25 (0.03) \\
        Gustavosta/MagicPrompt-Stable-Diffusion & 0.45 (0.23) & 0.29 (0.05) \\
        \bottomrule
    \end{tabular}}
\end{wraptable}

\textbf{Discussion}. Table \ref{tab:prompt-robust} presents two key robustness metrics for each LLM: the probability of statistically significant changes (P(p-value < 0.05)) and the effect size ($\omega$). Lower values in both metrics indicate higher robustness to irrelevant prompt perturbations. The results demonstrate a clear trend: more advanced models exhibit greater robustness. GPT-4 and GPT-3.5 show the lowest probabilities of significant changes (0.15 and 0.05) and among the smallest effect sizes (both 0.22-0.23). In contrast, smaller or less advanced models like SmolLM-135M and MagicPrompt-Stable-Diffusion show higher probabilities of significant changes (0.35 and 0.45) and larger effect sizes (0.25 and 0.29). This pattern aligns with the expectation that more sophisticated models should be less sensitive to irrelevant prompt variations, and the consistency across both metrics strengthens the reliability of these findings.

\subsection{DBPA can evaluate alignment with reference language model}
\label{sec:alignment}

\begin{wraptable}[14]{r}{.48\linewidth}
    \vspace{-\baselineskip-1.8pt}
    \centering
    \caption{\textbf{DBPA evaluate alignment with reference LLM}. The distance metric $\omega$ represents how far the LLM is from the reference, and the p-value shows whether that difference is statistically significant. As we expect, models trained on different datasets (e.g. Llama) are much different from GPT-4 compared to other models.}
    \vspace{-\baselineskip}
    \smallskip
    \label{tab:alignment}
    \resizebox{\linewidth}{!}{%
    \begin{tabular}{@{}l|c|c@{}}
        \toprule
        \bf LLM & \bf\makecell{$\omega$} & \bf\makecell{p-value} \\
        \midrule
        meta-llama/Meta-Llama-3.1-8B-Instruct & 0.31 (0.07) & 0.02 (0.02) \\
        HuggingFaceTB/SmolLM-135M & 0.29 (0.04) & 0.02 (0.04) \\
        openai-community/gpt2 & 0.28 (0.05) & 0.10 (0.13) \\
        gpt-35-1106 & 0.25 (0.05) & 0.24 (0.25) \\
        Gustavosta/MagicPrompt-Stable-Diffusion & 0.23 (0.06) & 0.37 (0.35) \\
        microsoft/Phi-3-mini-4k-instruct & 0.23 (0.04) & 0.28 (0.25) \\
        mistralai/Mistral-7B-Instruct-v0.2 & 0.21 (0.02) & 0.50 (0.21) \\
        google/gemma-2-9b-it & 0.19 (0.05) & 0.67 (0.31) \\
        \bottomrule
    \end{tabular}}
\end{wraptable}

One key flexibility of the DBPA framework is its ability to benchmark model perturbations in addition to input perturbations (which we have shown in earlier sections). When the model is perturbed rather than the input, what we are essentially measuring is the distance between answers from different models. Interestingly, this could act as a metric for alignment between models. 

\textbf{Setup}. This experiment demonstrates how the DBPA framework can evaluate alignment between different language models. We use GPT-4 (specifically gpt-4-0613-20231016) as the reference model. For a set of standardized questions, we first obtain answer distributions from GPT-4. We compute $\omega$ and the p-value.

\textbf{Discussion}. The results in Table \ref{tab:alignment} reveal varying degrees of alignment between different language models and GPT-4. The results demonstrate that the DBPA framework can quantify differences in response distributions between models, providing a concrete measure of inter-model alignment that could be useful for comparative analysis of language model outputs.

\section{Related work}
\label{sec:related_work}

There are three primary approaches to quantifying text-based outputs that relate to our approach. They include measuring unintended biases in model outputs, developing counterfacual fairness methods, and text summarization metrics.

$\blacktriangleright$ \textbf{Measuring unintended bias}. Overall, the closest related works are in measuring unintended bias \cite{borkan2019nuanced, dixon2018measuring, park2018reducing}. Such metrics quantify existing biases between subgroups for models. Broader work in the field can be found in \cite{friedler2021possibility, kleinberg2016inherent, menon2018cost}. However, this requires human annotation, relates \textit{only} to fairness, and assumes the existence of reliable labels across subgroups. 

$\blacktriangleright$ \textbf{Counterfactual fairness}. This approach \cite{garg2019counterfactual} examines how predictions would change if sensitive attributes were different. It can compute effect sizes but cannot be applied to black-box models, doesn't allow arbitrary perturbations, and doesn't enable statistical inference. It requires human input for labeling a specific attribute (e.g. toxicit) of an answer and makes assumptions, e.g. that non-toxic examples are less likely to contain asymmetric counterfactuals relative to toxic examples. 

$\blacktriangleright$ \textbf{Text summarization metrics}. These metrics \cite{bhandari2020re, zhang2019bertscore, zhao2019moverscore, lin2004rouge} evaluate the quality of text summarization. They can compute effect sizes but are not applicable to black-box models, don't allow arbitrary perturbations, and don't enable statistical inference. They require human input and make certain assumptions. Various metrics like BERTScore, MoverScore, and ROUGE variants measure different aspects of similarity between system and reference summaries.

To better explain how we differ, we compare each area based on five important criteria: (i) whether the method can be applied to any black-box model; (ii) whether any perturbation can be applied and measured; (iii) whether the approach enables statistical inference; (iv) whether the approach allows to compute effect sizes of the change; (v) whether there are any assumptions; (vi) whether humans are required as a part of the input. We show this in Table \ref{tab:llm_evaluation_methods}.

\begin{table}[htbp]
\centering
\tiny
\begin{tabular}{p{3.4cm}p{1.5cm}cccccp{3.5cm}}
\toprule
\textbf{Method} & \textbf{Example Works} & \textbf{(I)} & \textbf{(II)} & \textbf{(III)} & \textbf{(IV)} & \textbf{(V)} & \textbf{Representative Question} \\
\midrule
Measuring Unintended Bias & \cite{borkan2019nuanced, dixon2018measuring, park2018reducing} & \cmark & \xmark & \cmark & \cmark & \xmark & Does this model have unintended biases in certain subgroups? \\
\addlinespace
Counterfactual Fairness & \cite{garg2019counterfactual} & \xmark & \xmark & \xmark & \cmark & \xmark & How would the prediction change if the sensitive attribute were different? \\
\addlinespace
Text Summarization & \cite{bhandari2020re, zhang2019bertscore, zhao2019moverscore, lin2004rouge} & \xmark & \xmark & \xmark & \cmark & \cmark & How well is this text summarized? \\
\hline
distribution-based perturbation analysis & This work & \cmark & \cmark & \cmark & \cmark & \cmark & Do the responses change if we change any input in the prompt? If so, how? \\
\bottomrule
\end{tabular}
\caption{Related metrics of quantifying text-based outputs. \textbf{Abbreviations:} \textbf{(I)}: Usable on any black-box model; \textbf{(II)}: Any perturbation can be applied; \textbf{(III)}: Enables statistical inference; \textbf{(IV)}: Computes the effect size; \textbf{(V)}: Assumption-free}
\label{tab:llm_evaluation_methods}
\end{table}

\section{Discussion}

With the growing need to evaluate LLM systems, we require statistics-based approaches to understand LLM outputs. The introduction of distribution-based perturbation analysis is important for our ability to perform such evaluations. Apart from direct applications in high-stakes areas, we see such applications useful in other domains where language models are used as parts of broader statistical systems \cite{xi2023rise, rauba2024context, rauba2024self, astorga2024partially}. The versatility of DBPA in handling any input perturbation without requiring access to model internals makes it particularly valuable for practitioners and researchers working with diverse LLM architectures and deployment scenarios.

\textbf{Limitations.} While DBPA effectively tackles challenges of computational intractability and interpretability in analyzing LLM output distributions, the choice of similarity metrics and distance measures plays a critical role and can significantly impact results. Future work should investigate these choices in depth and provide clear guidelines for selecting appropriate measures for specific contexts. Additionally, although DBPA offers valuable insights for model auditing, translating these findings into practical strategies for enhancing model robustness and aligning outputs with human preferences remains a significant challenge. Exploring uncertainty-aware methods, such as those in \cite{liu2024large}, to identify perturbations that achieve specific objectives presents a promising research direction. Bridging the gap between identifying sensitivities and implementing targeted interventions is essential for advancing LLM performance and reliability.

\newpage
%%%%%%%%%%%%%%%%%%%%%%%%%%%%%%%%%%%%%%%%%%%%%%%%%%%%%%%%%%%%
\bibliographystyle{unsrt}
\bibliography{export} 

\newpage
\appendix

\section{MC-Sampling}
\label{appendix:mc_sampling}

To understand why we employ monte-carlo sampling, let's consider the expectation of $\omega$ under both approaches. For permutation testing, this yields:

\begin{equation}
    \mathbb{E}[\omega(P_0^*, P_1^*)] = \mathbb{E}[\omega(\{s(z_i, z_j)\}, \{s(z_i, z_k)\})] \quad z_i, z_j, z_j \stackrel{i.i.d.}{\sim} \mathcal{Z}
\end{equation}

where $\mathcal{Z} = \{y_i, \ldots, y_k, y'_1, \ldots, y'_k\}$ is the combined set of 2k samples from $S(x)$ and $S(\Delta(x))$. For resampling, the expectation is:

\begin{equation}
    \mathbb{E}[\omega(P'_0, P'_1] = \mathbb{E}[\omega(\{s(y_i, y_j)\}, \{(y_i, y'_k)\})] \quad y_i, y_j \stackrel{i.i.d.}{\sim}\mathcal{S}(x), y'_k \stackrel{i.i.d.}{\sim} \mathcal{S}(\Delta(x))
\end{equation}

Under $H_0$, $\mathcal{S}(x) = \mathcal{S}(\Delta(x))$, so all samples in $\mathcal{Z}$ follow the same distribution.  Therefore, randomly sampling from $\mathcal{Z}$ is equivalent to sampling from either $\mathcal{S}(x)$ or $\mathcal{S}(\Delta(x))$. Therefore, the expectations are identical, and the permutation testing will produce the same distribution of $\omega$ values in the limit.

\section{Examples of prompt inputs}
\subsection{Example inputs for treatment recommendations}

In the exercise for treatment recommendations, we use the following input prompts (where characteristics can vary):

John has these features: Age: 70, BMI: 30, Blood Pressure: 114/61 mmHg, 
    Total Cholesterol: 6.7 mmol/L, HDL Cholesterol: 1.0 mmol/L, 
    Smoking Status: current smoker, Diabetes: no diabetes, 
    Family History of CVD: yes, Ethnicity: Chinese. Based on these characteristics alone, provide recommendations on CVD guidelines based on NICE for this person

John has these features: Age: 70, BMI: 29, Blood Pressure: 114/61 mmHg, 
    Total Cholesterol: 6.7 mmol/L, HDL Cholesterol: 1.0 mmol/L, 
    Smoking Status: current smoker, Diabetes: no diabetes, 
    Family History of CVD: yes, Ethnicity: Chinese. Based on these characteristics alone, provide recommendations on CVD guidelines based on NICE for this person

\subsection{Example outputs for treatment recommendations}

The following are example outputs:

\begin{lstlisting}
    ["Based on these characteristics, NICE guidelines for CVD recommend the following:\n\n1. Encouraging John to quit smoking and providing support and resources to help him do so.\n2. Promoting a healthy diet and regular physical activity to help lower his BMI and improve his cholesterol levels.\n3. Monitoring and managing his blood pressure to keep it within a healthy range.\n4. Considering his family history of CVD, it may be advisable to assess his overall cardiovascular risk and consider interventions to reduce that risk.\n5. Providing culturally sensitive and tailored support for his Chinese ethnicity, taking into account any specific risk factors or considerations.\n\nIt's important for John to work with his healthcare provider to develop a personalized plan for cardiovascular health and to address any individualized risk factors or concerns.",
  "Based on the provided characteristics, John's CVD risk would be assessed as moderate to high. The NICE guidelines recommend the following for individuals with similar characteristics:\n\n1. Lifestyle changes: John should be advised to make lifestyle changes such as quitting smoking, following a healthy diet to reduce cholesterol levels and maintain a healthy weight, and engage in regular physical activity.\n\n2. Assess CVD risk: John should undergo a formal assessment of CVD risk using a cardiovascular risk assessment tool, taking into account his age, BMI, blood pressure, cholesterol levels, smoking status, and family history of CVD to determine the need for medication or further interventions.\n\n3. Regular monitoring: John should have regular monitoring of his blood pressure, cholesterol levels, and overall cardiovascular health to manage and reduce his risk of CVD.\n\n4. Consider medication: Depending on the results of his CVD risk assessment, John may need to be considered for medication such as statins to lower his cholesterol levels and reduce his overall CVD risk.\n\nIt's important for John to work closely with a healthcare professional to develop a personalized plan for managing his CVD risk based on the NICE guidelines.",
  "Based on the provided characteristics, NICE guidelines for CVD prevention would recommend the following:\n\n1. Encouraging John to quit smoking and providing support and resources to help him quit.\n2. Recommending lifestyle modifications such as regular physical activity and a healthy diet to help manage his BMI and cholesterol levels.\n3. Monitoring and managing his blood pressure to keep it within a healthy range.\n4. Consideration of cholesterol-lowering medication if lifestyle modifications are not sufficient to achieve target cholesterol levels.\n5. Regular monitoring and management of other CVD risk factors, especially considering his family history of CVD.\n\nIt's important for John to work closely with his healthcare provider to create a personalized plan for CVD prevention and management based on his individual risk factors and health needs.",
  "Based on the provided characteristics, the following recommendations can be made for CVD guidelines based on NICE:\n\n1. Lifestyle changes: John should be advised to quit smoking and adopt a healthier lifestyle, including regular physical activity and a balanced diet to help lower his BMI and cholesterol levels.\n\n2. Cardiovascular risk assessment: John's overall cardiovascular risk should be assessed using a validated tool, taking into account his age, blood pressure, and cholesterol levels.\n\n3. Medical assessment: John should be assessed for any potential risk factors for cardiovascular disease, such as high blood pressure, and be offered appropriate management and treatment as necessary.\n\n4. Family history: Given John's family history of CVD, he should be counseled on the potential genetic predisposition to cardiovascular disease and the importance of managing other risk factors effectively.\n\n5. Ethnicity: Being of Chinese ethnicity, John should be aware that certain populations may have a higher risk for specific cardiovascular conditions, and this should be taken into consideration when assessing his overall cardiovascular risk.\n\nIt is important to note that these recommendations are general and may need to be tailored specifically to John's individual health needs and circumstances. It is advisable for John to seek personalized medical advice and undergo a comprehensive cardiovascular risk assessment and management plan under the care of a healthcare professional.",
\end{lstlisting}

%%%%%%%%%%%%%%%%%%%%%%%%%%%%%%%%%%%%%%%%%%%%%%%%%%%%%%%%%%%%

\end{document}